\documentclass[runningheads,a4paper]{llncs}

\usepackage{amssymb}
\usepackage{graphicx}
\usepackage{float}
\usepackage[caption = false]{subfig}
\usepackage{multirow}
\usepackage{url}
\usepackage{amsmath}

\urldef{\mailsa}\path|{alfred.hofmann, ursula.barth, ingrid.haas, frank.holzwarth,|
	\urldef{\mailsb}\path|anna.kramer, leonie.kunz, christine.reiss, nicole.sator,|
	\urldef{\mailsc}\path|erika.siebert-cole, peter.strasser, lncs}@springer.com|

\begin{document}
	
	\mainmatter  % start of an individual contribution
	
	\title{MSCE: An edge preserving robust loss function for improving super-resolution algorithms}
	
	\titlerunning{ }
	
	\author{Ram Krishna Pandey \and Nabagata Saha\and Samarjit Karmakar\and A G Ramakrishnan}
	\authorrunning{Ram Krishna Pandey et al.}
	
	\institute{Indian Institute of Science,\\
		Bangalore, India\\
		%\url{http://www.springer.com/lncs}
	}
	
	
	%\toctitle{Lecture Notes in Computer Science}
	
	%\tocauthor{Authors' Instructions}
	
	\maketitle
	
	\begin{abstract}
		
		With the recent advancement in the deep learning technologies such as CNNs and GANs, there is significant improvement in the quality of the images reconstructed by deep learning based super-resolution (SR) techniques. In this work, we propose a robust loss function based on the preservation of edges obtained by the Canny operator. This loss function, when combined with the existing loss function such as mean square error (MSE), gives better SR reconstruction measured in terms of PSNR and SSIM. Our proposed loss function guarantees improved performance on any existing algorithm using MSE loss function, without any increase in the computational complexity during testing.
	
	\end{abstract}

	\section{Introduction}
	Super-resolution is the process of obtaining a high resolution (HR) image from one or more low resolution (LR) images. Classical reconstruction based image super-resolution requires multiple low-resolution images with sub-pixel misalignment at the same scale, whereas single image super-resolution requires a database of LR and HR matched pairs to learn a mapping function between the patch pairs at different scales~\cite{glasner}. Given a low resolution image during testing, this learned function or representation can be used to reconstruct the corresponding HR image.
	
	Since the advent of deep learning technologies in the past decade, super-resolution algorithms have shown remarkable improvement in the quality of the reconstructed image. Most of the work reported in the literature have used mean square error (MSE) loss function to minimize the error between the reconstructed model output and the ground truth image~\cite{srcnn14}~\cite{subpixel}~\cite{rkagrnatural}~\cite{rkagrdocument1}~\cite{rkagrdocument2}~\cite{rkagrdocument3}. Minimizing this loss function may reduce the high frequency content in the reconstructed image and thus may blur the edges in it. Also, the reconstructed image may not lie precisely in the manifold of the HR image. Researchers have endeavored to find ways to solve this problem to a good extent as can be seen in SRGAN~\cite{srgan}, where the authors claim that the reconstructed output lies precisely in the manifold of HR images, even if the reconstructed images have less peak signal to noise ratio (PSNR) and structural similarity (SSIM). Ledig et al. \cite{srgan} have used a weighted combination of MSE loss, content loss~\cite{perceptualloss} and adversarial loss to reconstruct the HR image. This approach requires a deep architecture, such as the VGG net~\cite{vggnet}, to obtain the local covariance structure in the image. Most of the image transformation tasks use mean square error as loss function, which provides smooth transformed images.  
	
	Our main contributions in this paper are as follows: 
	\begin{itemize}
		\item We have performed a large number of experiments to obtain a robust loss function that improves the performance of the existing algorithms that employ MSE loss function.

		\item While training, we apply Canny edge detector \cite{canny} the reconstructed output (in batches) and also separately on the corresponding ground truth image to compute the proposed mean square Canny error (MSCE) and assign weights (convex combination) based on our experiments i.e. the loss function can be given as: $ \mu \times MSE + (1-\mu)\times MSCE $.
		
		\item Our approach guarantees performance improvement in terms of PSNR and SSIM over the existing approaches, if the model is trained on one dataset and tested on different datasets as mentioned in Tables~\ref{result_table_all1} and~\ref{result_table_all2}.
		
		\item  Our model does not incur additional overhead in terms of computation during testing to obtain the performance gain reported in Tables~\ref{result_table_all1} and~\ref{result_table_all2} due to our proposed MSCE loss function.
		
	\end{itemize}
	
	\section{Related work}
	Super-resolution and image denoising can be assumed as image transformation tasks. In super-resolution, a LR image is fed to a transformation network such as a multilayer neural network to generate a HR image. Most of the image processing tasks such as image denoising and super-resolution minimize a per-pixel loss function to obtain reconstruction. In this work, our focus is on improving the quality of existing super-resolution algorithms such as SRCNN~\cite{srcnn14} and ESPCN~\cite{subpixel} that use per-pixel loss function. Recently proposed perceptual loss function has shown significant improvement in the perceptual quality of the images. Simonyan et al~\cite{featurevisualization} use perceptual loss for feature visualization. Gatys et al.~\cite{Gatystexture} and \cite{Gatysstyle} use perceptual loss for texture synthesis and style transfer, respectively. These approaches solve an optimization problem and hence, are slower.

	Justin Johnson et al~\cite{Johnson} and Pandey et al~\cite{rkagrart} use the benefits of per-pixel as well as perceptual loss funtions and propose a computationally efficient, optimization-free approach that provides results for image transformation tasks that are qualitatively similar to those of the above optimization-based approaches. The super-resolution algorithm SRGAN~\cite{srgan} uses a weighted combination of three different loss functions, namely mean square error, perceptual and adversarial loss to obtain a sharper reconstruction. The images reconstructed by these methods perceptually look sharper, even if they have low values of PSNR and SSIM.

	In this work, our focus is on improving the perceptual quality, PSNR and SSIM without incurring any additional computational overhead during testing by the addition of a new, robust loss function that aims to preserve the edge information.

	\section{The proposed edge-preserving MSCE loss function}
	We employ Canny edge detector~\cite{canny} to detect the edges in the reconstructed and ground truth images. We have chosen this algorithm, since Canny operator provides the most reliable edges amongst all the edge detection algorithms in the literature, and also satisfies all the general edge detection criteria. 
	
	Most of the recent papers on image super-resolution and denoising use mean square error as the loss function. This loss function may smooth the edge components in an image. We thought of preserving the edges by defining the loss function as a convex combination of mean square error loss and our edge preserving loss as follows:
	
	Suppose the training set consists of image pairs $\{L_{i},H_{i}\}$ ; $ i = 1... N $, where N is the total number of training examples. The model $\Theta$, parameterized by $ \lambda$, predicts the output $ O_{j}$ for a given input $L_{j}$. Let C denote the Canny operator. Let $C(\Theta_{\lambda}(L_{j})))$ be the resultant image obtained by applying Canny operator on the predicted output image, $O_{j} = \Theta_{\lambda}(L_{j})$. The proposed edge preserving loss function, called the mean square Canny error - (MSCE) is given by :
	
	\begin{equation}
	\scriptsize{
		Loss  =\underbrace{\mu\times\frac{1}{N}\sum\limits_{j=1}^{N}\parallel \Theta_{\lambda}(L_{j})-H_{j}\parallel_{F}}_{MSE \hspace{0.1cm}Loss\hspace{0.1cm}(l_{mse})} + \underbrace{(1-\mu)\times\frac{1}{N}\sum\limits_{j=1}^{N}\parallel C(\Theta_{\lambda}(L_{j}))-C(H_{j})\parallel_{F}}_{Edge\hspace{0.1cm}preserving\hspace{0.1cm} loss\hspace{0.1cm}(l_{edge})}}
	\label{equation1}
	\end{equation}

	The first term in the equation above is the mean square loss function used to minimize the error between the reconstructed output and the ground truth image. The second term in the loss function is the edge preserving loss function. After a large number of experiments, the weighing factor $\mu$ has been fixed to lie in the range $0.8\leq \mu \leq 0.99$. To minimize this loss function, Adam optimizer~\cite{adam} is used with learning rate $(lr) =0.001$, $\beta_{1}=0.999$ and $\beta_{2}=0.99$.

	\subsection{Choosing the value of $\mu$}
	
	\begin{itemize}
	\item {\bf Exhaustive Experimentation} : We performed experiments varying $\mu $ in the range $0.8 \leq \mu \leq 0.99$ by incrementing its value by 0.01 each time. We found that the models were consistently giving better results for the particular values of $\mu$ = 0.84, 0.85 and 0.86. For the results reported in the Figures~\ref{Fig1},~\ref{Fig2},~\ref{Fig3} and~\ref{Fig4} and Tables~\ref{result_table_all1} and ~\ref{result_table_all2}, the value of $\mu$ used is 0.85. \\
		
	\item {\bf Dynamic choice of $\mu$}: While performing the experiments, we found that sometimes, values of $\mu$ (still in the range $0.8\leq \mu \leq 0.99$) other than the three specific ones mentioned above, gave better results. We made a list of those different values of $\mu $ and tried each of them parallely in each epoch. For the subsequent epoch, we select the model corresponding to the least value of the loss function. Let $l_{mse}$ and $l_{edge}$ denote the mean square error loss and our edge-preserving loss, respectively, as mentioned in equation~\ref{equation1}. Let $ \{ \hat{\lambda} ,\hat{ \mu\ } \} $ be the optimal model parameters and $\mu$ be the weighing parameter currently chosen during training. In each epoch, we selected the value of $\mu $ that minimized the loss function in the right hand side of equation~\ref{equation2}:
		\begin{equation}
		\hat{ \mu\ } = \underset{\mu}{\operatorname{argmin}}\{\mu \times l_{mse}+ (1-\mu)\times l_{edge}\}
		\label{equation2}
		\end{equation}

	We used the earlier approach for calculating the loss in our experiments, results for which have been reported in the Tables. A dynamic choice of the value of $\mu$ gives similar results in less number of epochs. It can be experimented further to possibly achieve still better results.

	\end{itemize}
\section{Datasets used for training and testing}
	
	The models are trained on DIV2K~\cite{div2k} training dataset with the original architecture (without changing the architectural details of the existing model) proposed in the respective papers. We have performed testing on different datasets such as Set5~\cite{set5}, Set14~\cite{set14}, BSD~\cite{bsd100} and URBAN100~\cite{urban100} for the different scale factors of 2, 3, 4 and 8. We have found that there is consistent performance gain over the original models, in terms of PSNR and SSIM, on all the datasets on which our MSCE loss function has been tested so far. These results are seen quantitatively in Tables~\ref{result_table_all1} and~\ref{result_table_all2}.
	
\section{Experiments and Discussion}
	
	We have performed extensive experiments on different super-resolution algorithms proposed recently, by augmenting the original loss function with our proposed mean square Canny error loss function.

	We have validated the effectiveness of our proposed MSCE loss function on the recent techniques of SRCNN~\cite{srcnn14} and ESPCN~\cite{subpixel}. We found that the addition of our MSCE loss leads to better results and the improvement is consistent on both methods across different upscaling factors of 2, 3, 4 and 8.

\section{Results}	
\begin{figure}
		\subfloat[][SRCNN original]{\includegraphics[width=3cm]{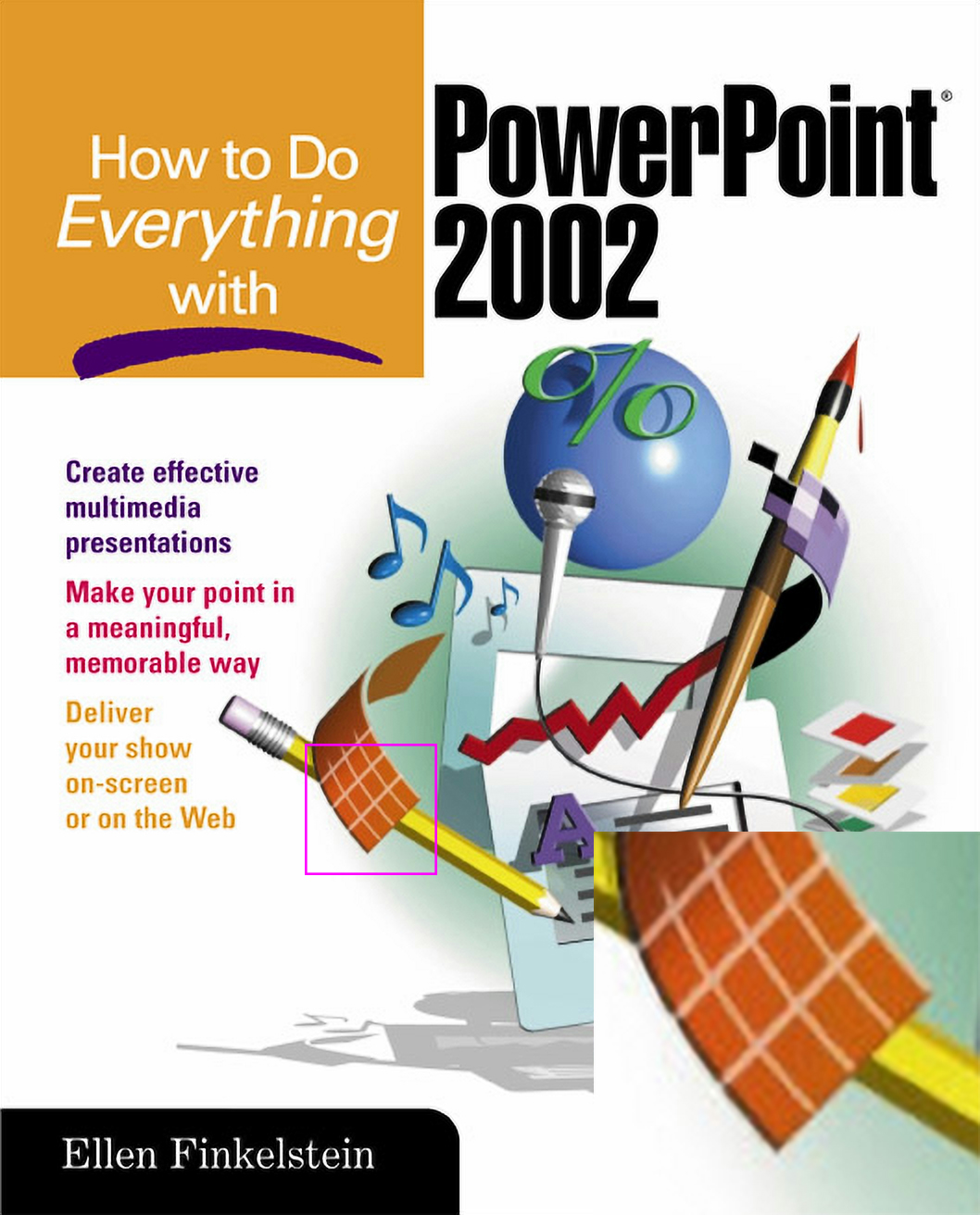}\label{ppt s-original}}
	\subfloat[][SRCNN MSCE]{\includegraphics[width=3cm]{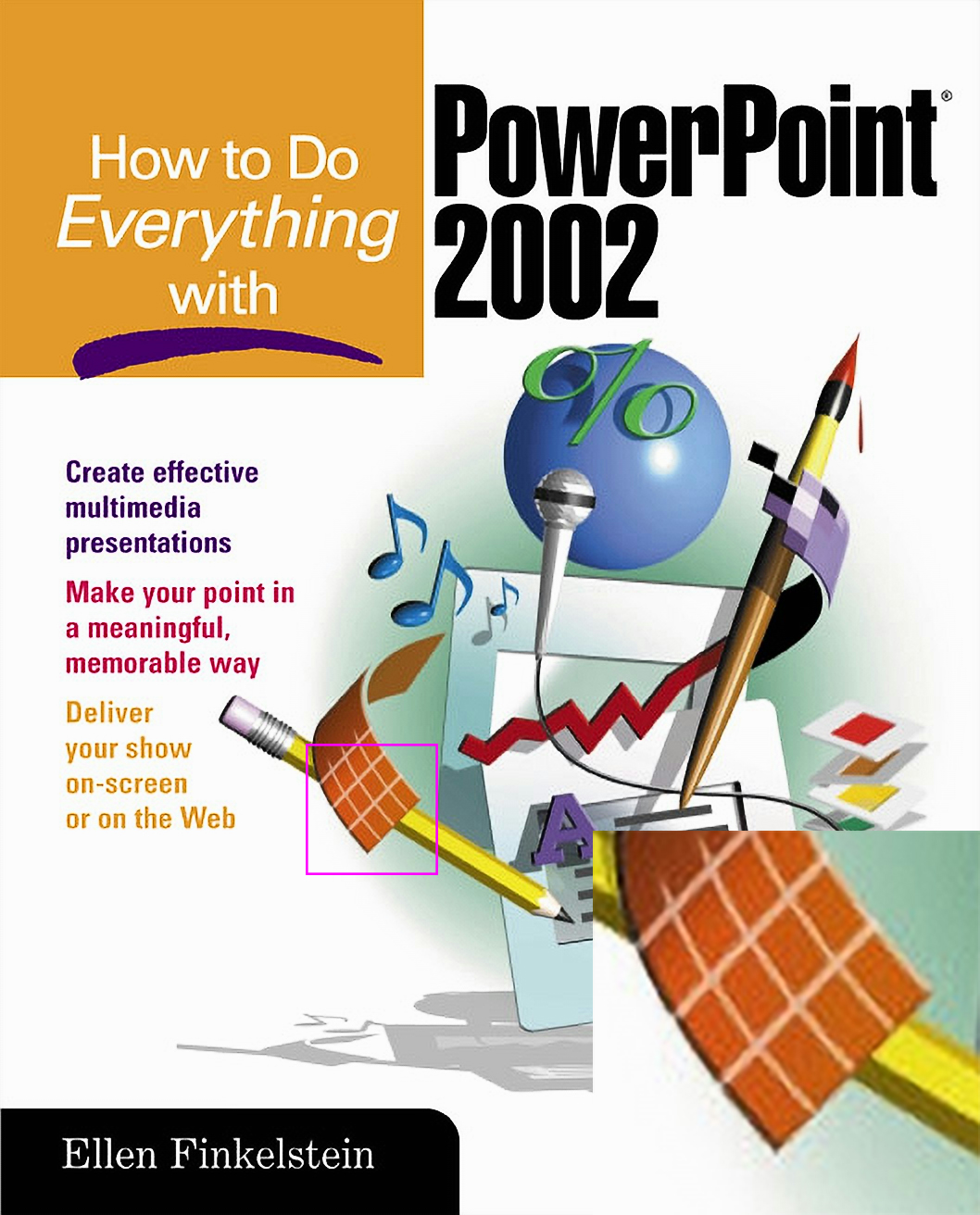}\label{ppt s-custom}}
	\subfloat[][ESPCN original]{\includegraphics[width=3cm]{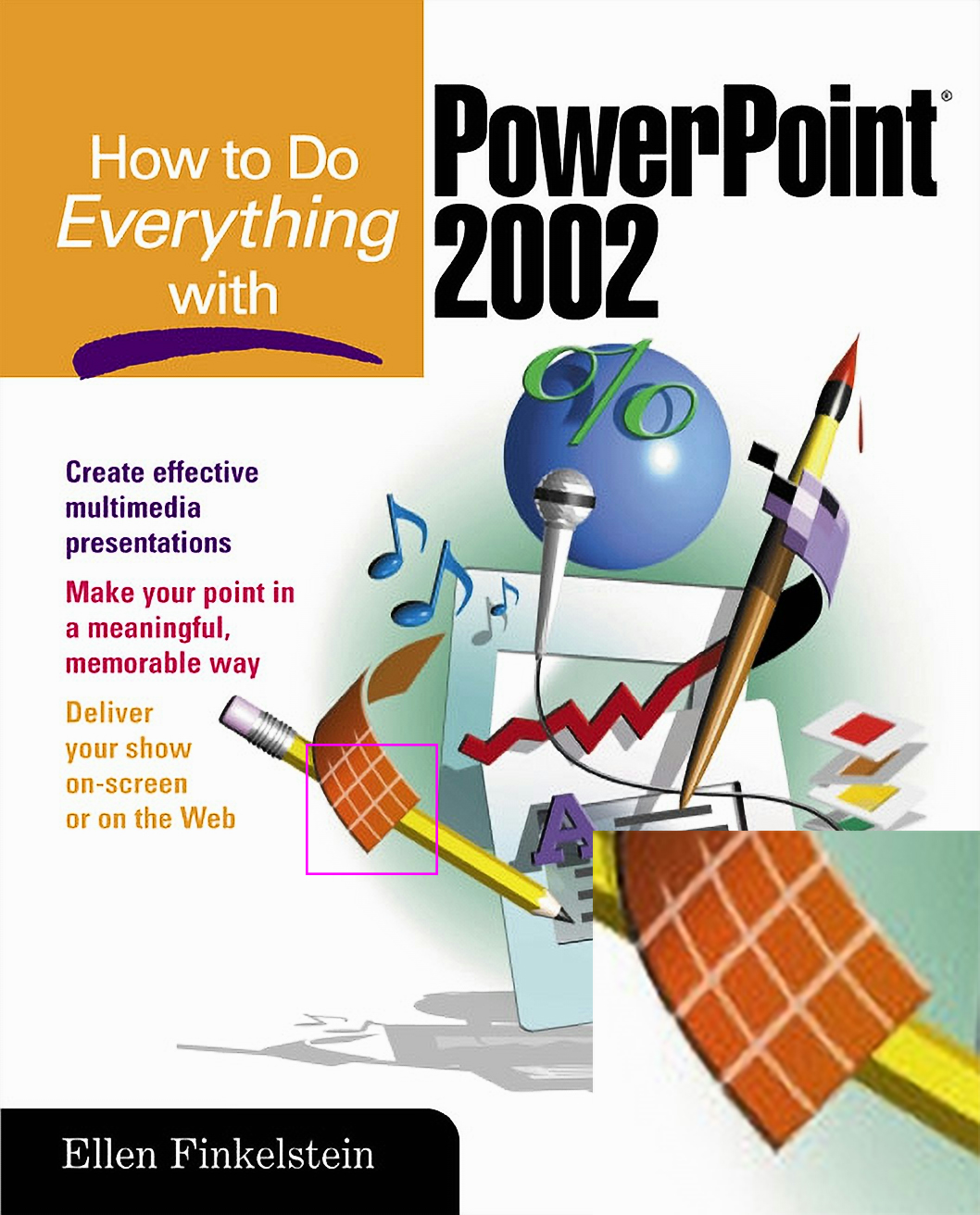}\label{ppt e-original}}
	\subfloat[][ESPCN MSCE]{\includegraphics[width=3cm]{ppt3_custom_espcn2}\label{ppt e-original}}
	\caption{Qualitative comparison of the results for an upscale factor of 2, when the ppt image from Set14 is directly fed to the original model and the model modified with MSCE loss trained by us. (a) The output image reconstructed by the original SRCNN model. (b) The output image reconstructed by SRCNN model modified with MSCE loss function. (c) Image reconstructed by the original ESPCN model. (d) Output image reconstructed by ESPCN model modified with MSCE loss function.}
	\label{Fig1}
	\end{figure}

\begin{figure}
	\centering
	\subfloat[][SRCNN original]{\includegraphics[width=3cm]{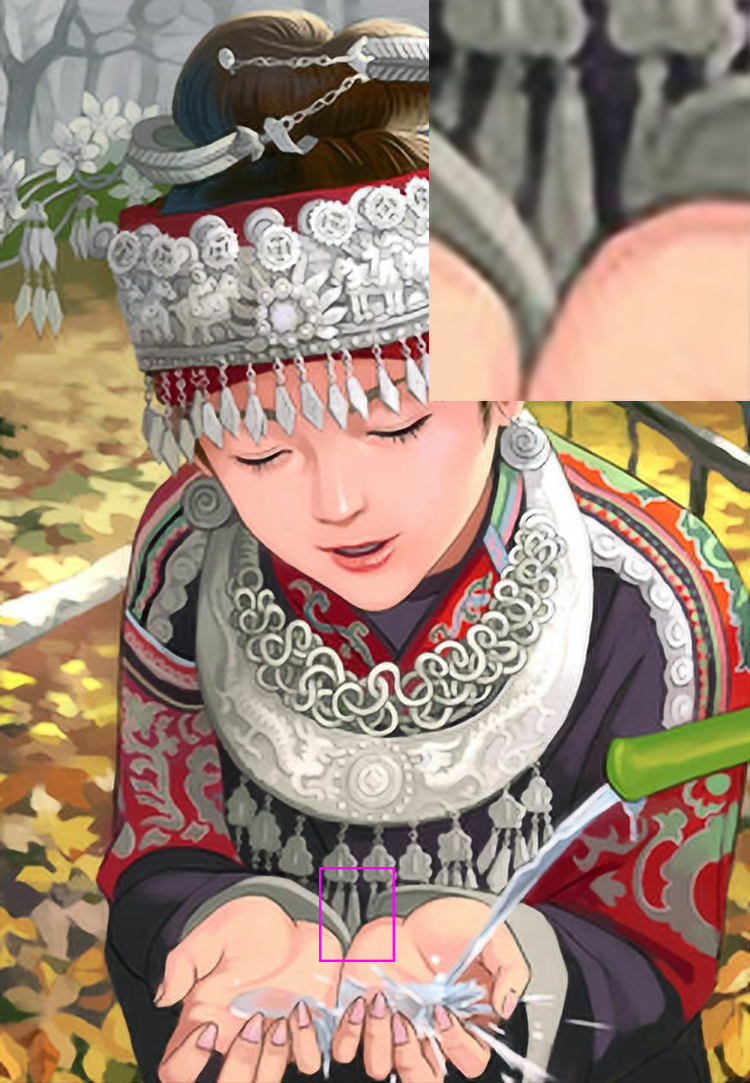}\label{comic original}}
	\subfloat[][SRCNN MSCE]{\includegraphics[width=3cm]{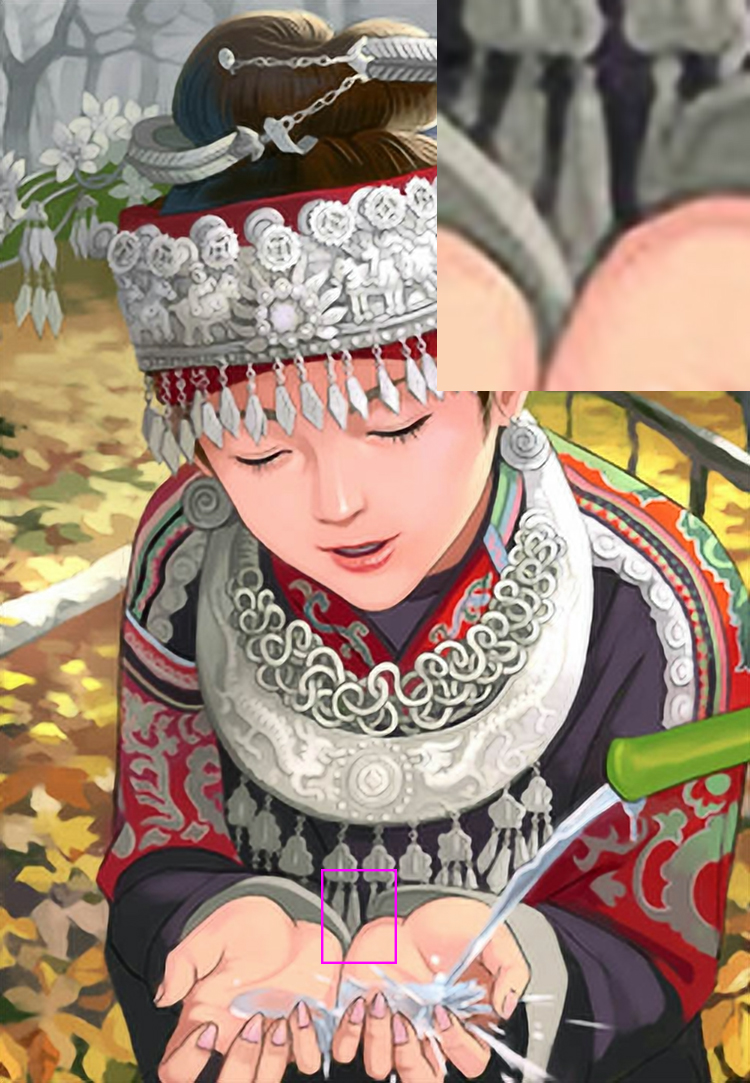}\label{comic custom}}
	\subfloat[][ESPCN original]{\includegraphics[width=3cm]{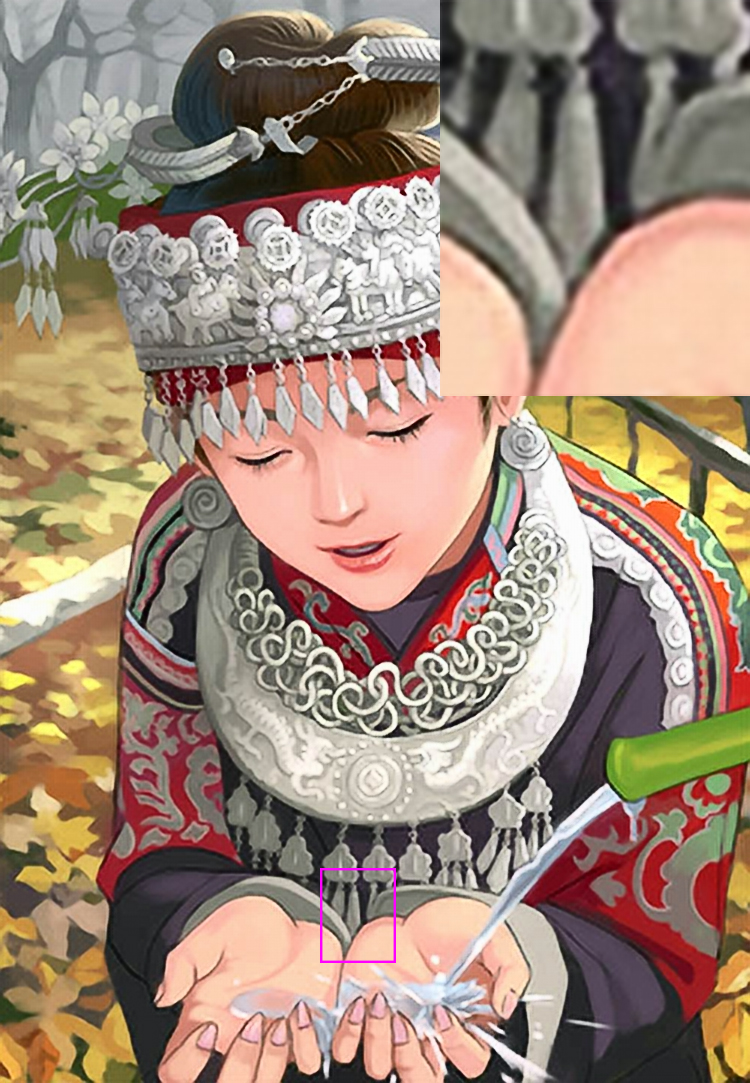}\label{comic custom}}
	\subfloat[][ESPCN MSCE]{\includegraphics[width=3cm]{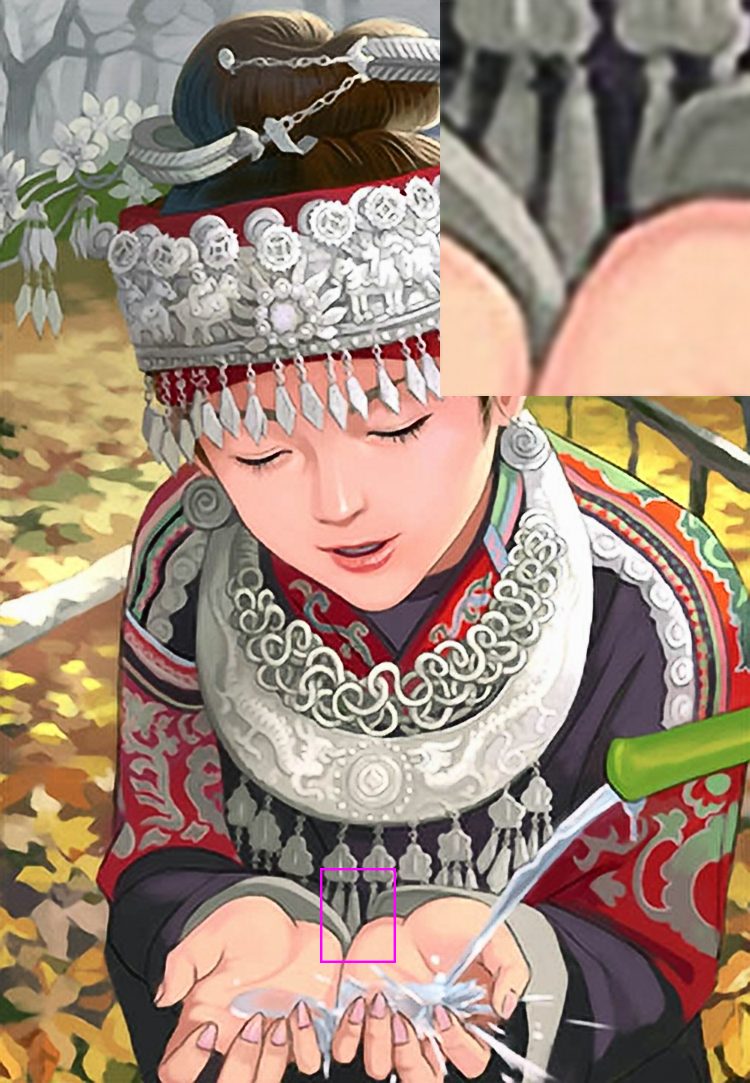}\label{comic custom}}
		
\caption{Comparison of the results for an upscale factor of 3, when the comic image from Set14 is directly fed to the original model and the model modified with MSCE loss trained by us. (a) Output image reconstructed by the original SRCNN model. (b) Output image reconstructed by SRCNN model modified by MSCE loss function. (c) Output image reconstructed by the original ESPCN model. (d) Output image reconstructed by ESPCN model modified by MSCE loss function.}
	\label{Fig2}
	\end{figure}

\begin{figure}
	\subfloat[][SRCNN original]{\includegraphics[width=3cm]{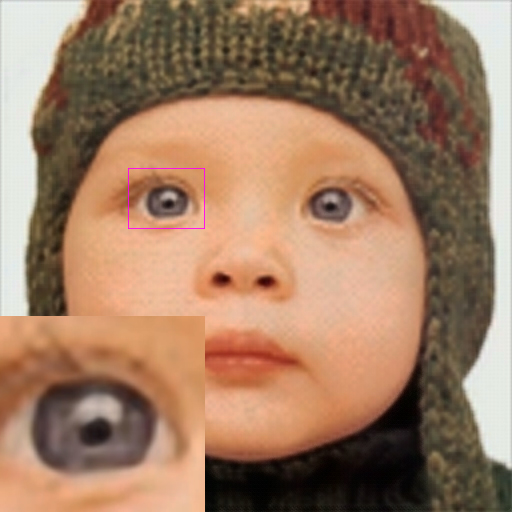}\label{baby original}}
	\subfloat[][SRCNN MSCE]{\includegraphics[width=3cm]{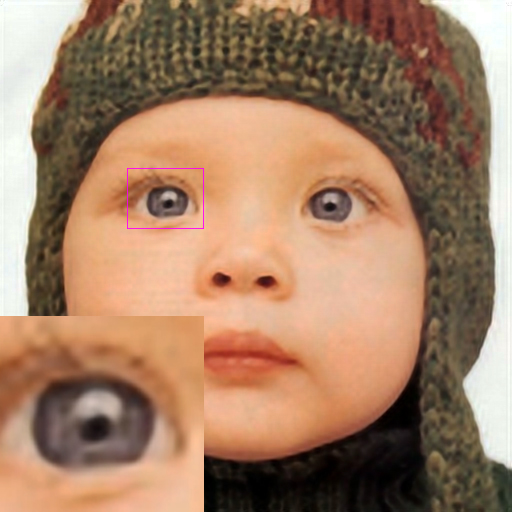}\label{baby custom}}
	\subfloat[][ESPCN original]{\includegraphics[width=3cm]{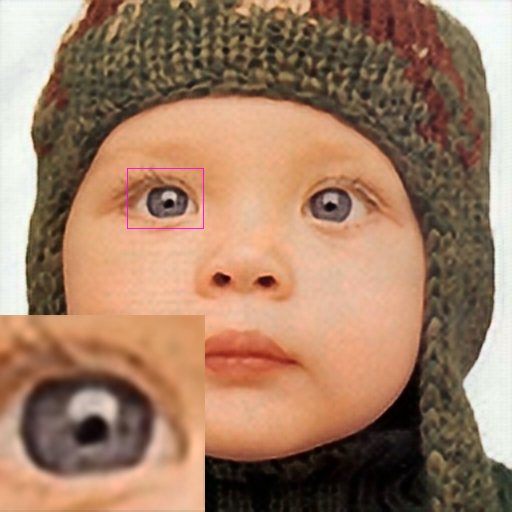}\label{baby original}}
	\subfloat[][ESPCN MSCE]{\includegraphics[width=3cm]{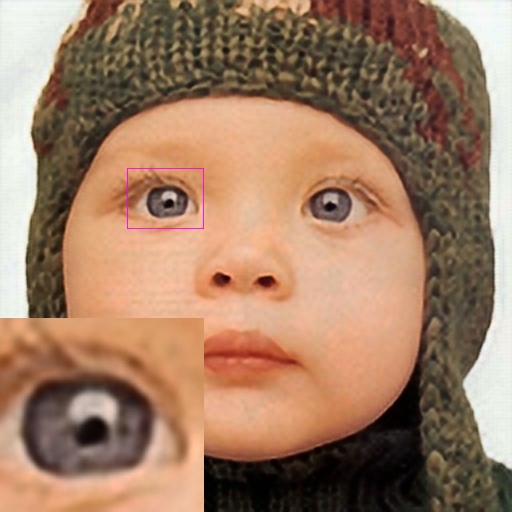}\label{baby original}}
	\caption{Comparison of the results for an upscale factor of 4, when the baby input image from Set5 is directly fed to the original model trained by us and the model modified with MSCE loss trained by us. (a) Output image reconstructed by the original SRCNN model. (b) Output image reconstructed by SRCNN model modified by MSCE loss function. (c) Output image reconstructed by the original ESPCN model. (d) Output image reconstructed by ESPCN model modified by MSCE loss function.}
	\label{Fig3}
	\end{figure}

\begin{figure}
	\subfloat[][SRCNN original]{\includegraphics[width=3cm]{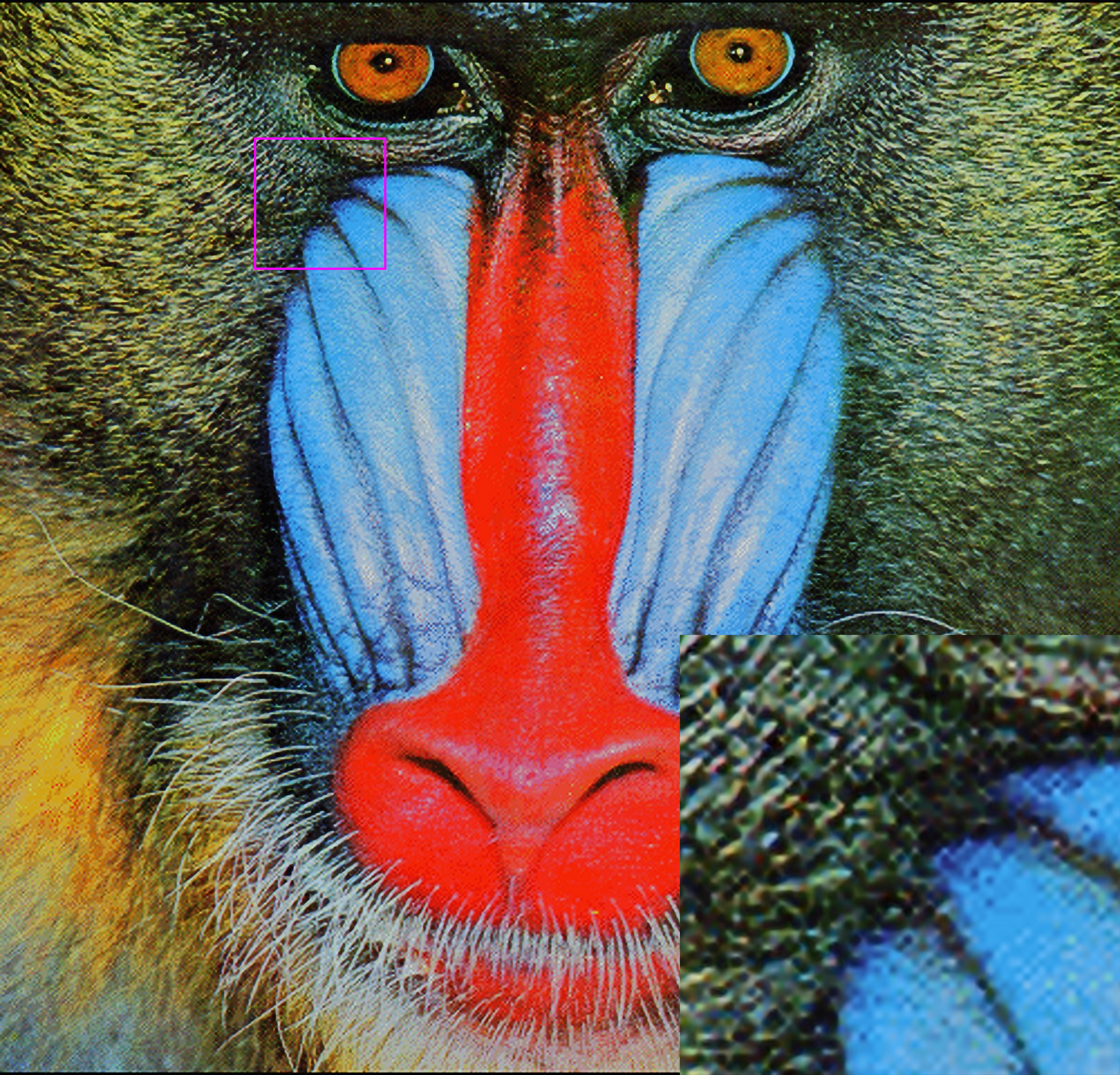}\label{baboon s-original}}
	\subfloat[][SRCNN MSCE]{\includegraphics[width=3cm]{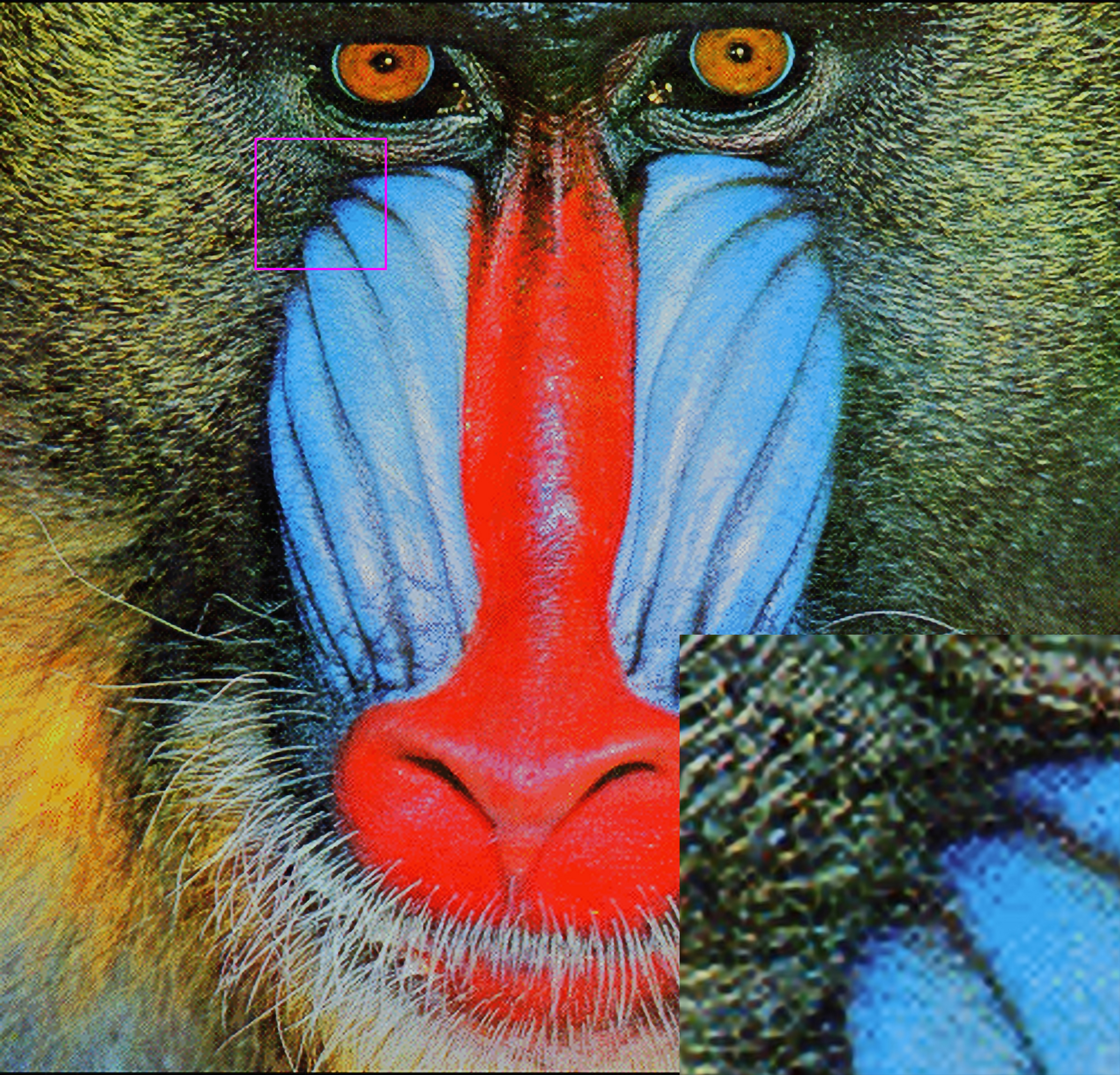}\label{baboon s-custom}}
	\subfloat[][ESPCN original]{\includegraphics[width=3cm]{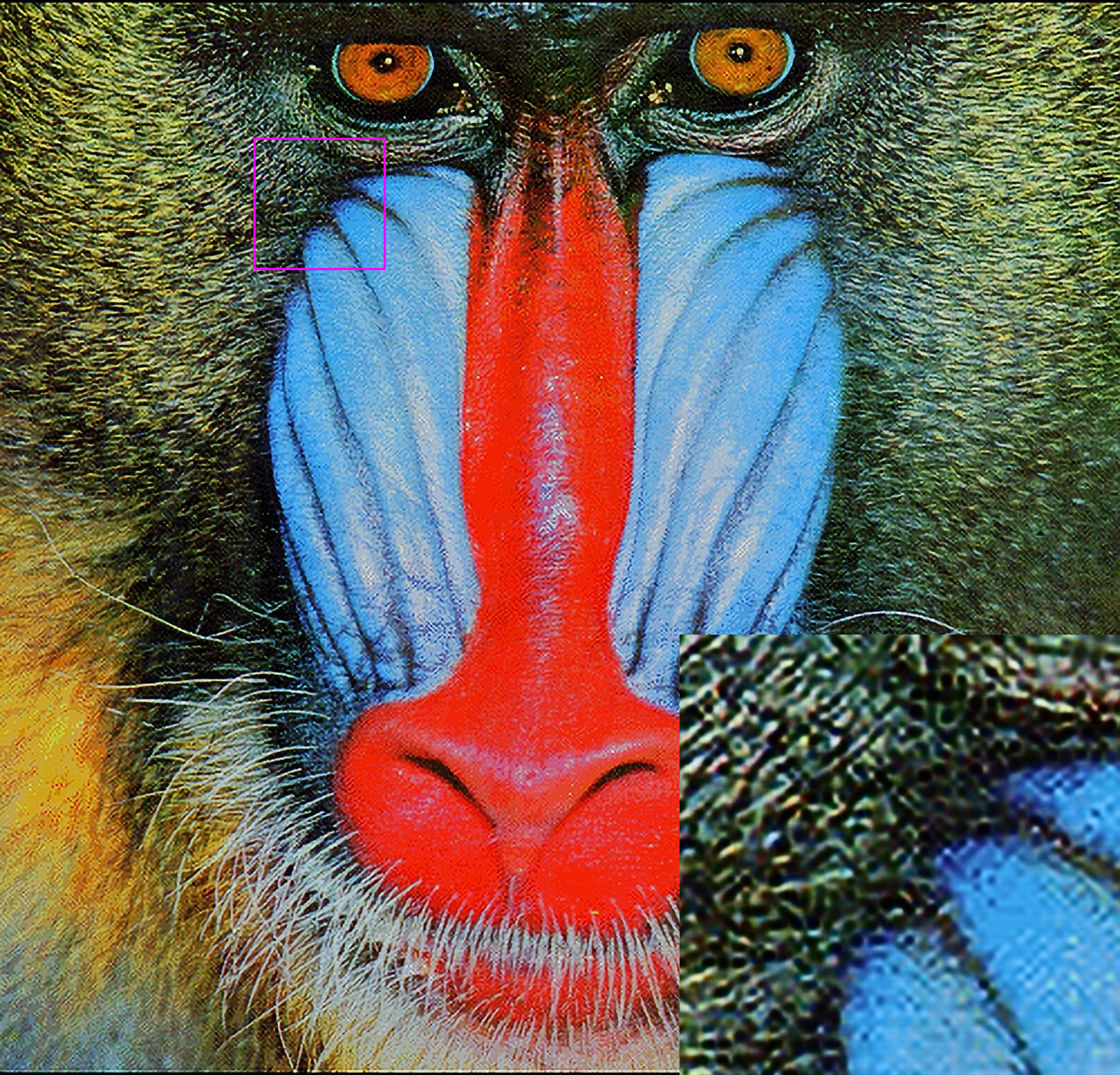}\label{baboon e-original}}
	\subfloat[][ESPCN MSCE]{\includegraphics[width=3cm]{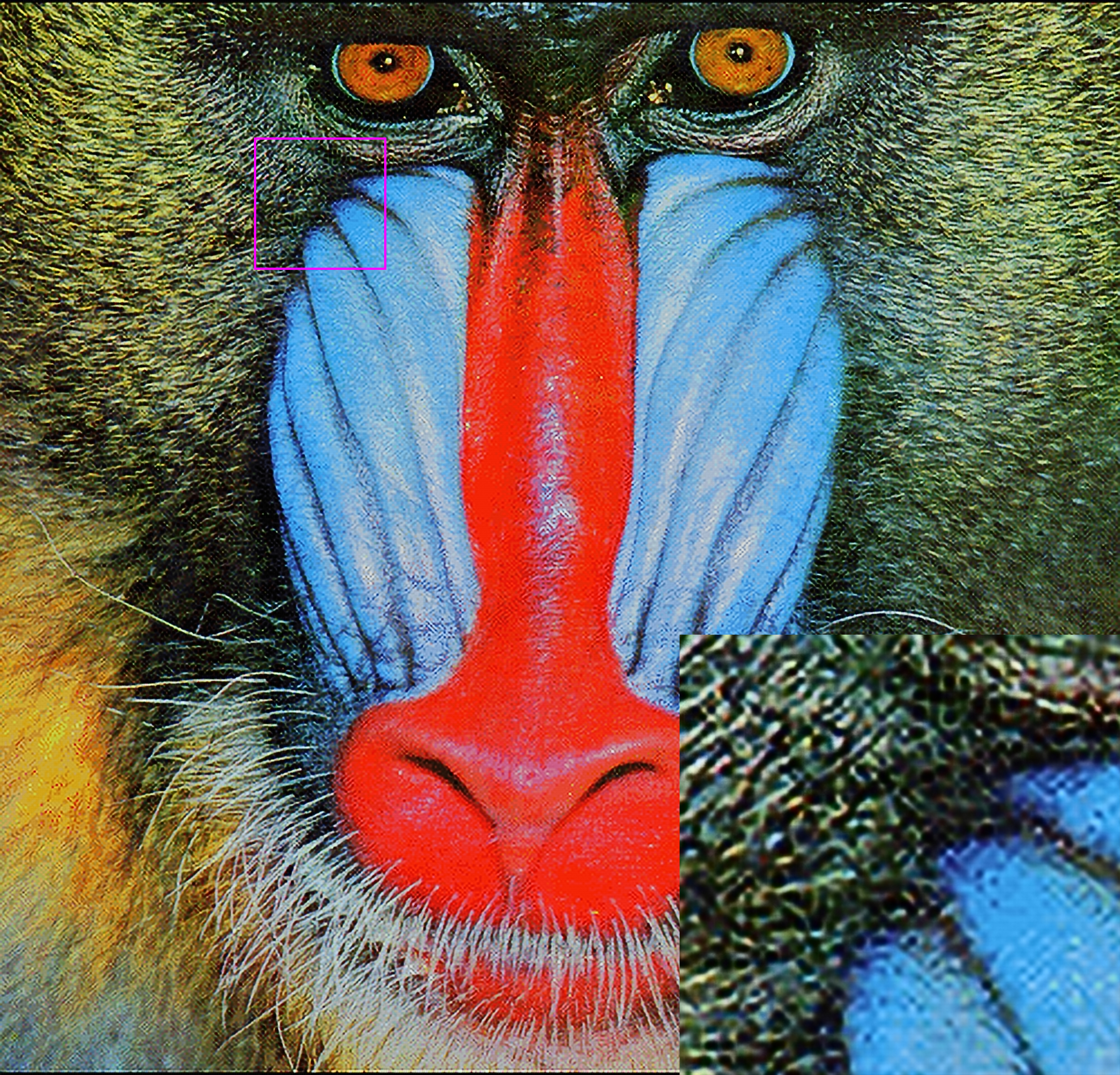}\label{baboon e-custom}}
	\caption{Comparison of the results for an upscale factor of 8, when the baboon input image from Set14 is directly fed to the original model trained by us and the model modified with MSCE loss trained by us. (a) Output image reconstructed by the original SRCNN model. (b) Output image reconstructed by SRCNN model modified by MSCE loss function. (c) Output image reconstructed by the original ESPCN model. (d) Output image reconstructed by ESPCN model modified by MSCE loss function.}
	\label{Fig4}
\end{figure}

	Figures~\ref{Fig1},~\ref{Fig2},~\ref{Fig3} and~\ref{Fig4} show both the results qualitatively: one obtained by passing the input image directly to the original models SRCNN~\cite{srcnn14} and ESPCN~\cite{subpixel} with the loss functions used in the original papers, and the other obtained by augmenting the loss function with our MSCE loss function.

	Tables~\ref{result_table_all1} and~\ref{result_table_all2} list the quantitative results obtained by the two superresolution methods on the datasets Set5, Set14, URBAN and BSD for different upscaling factors and the corresponding values obtained after they are modified by our MSCE loss function.

	{\bf Note 1:} Table~\ref{result_table_all1} lists the results obtained from the LR images created by downsampling using normal bicubic interpolation. Whereas, the results reported in Table~\ref{result_table_all2} are obtained by blurring the image by a Gaussian filter with radius 2 and then downsampling by bicubic interpolation to obtain the LR images at different scales.

	%{\bf Note 2:} We have not performed center cropping to remove boundary artifacts while computing PSNR and SSIM.
	

\begin{figure}
	\subfloat[][SRCNN   2x] {\includegraphics[width=3.1cm]{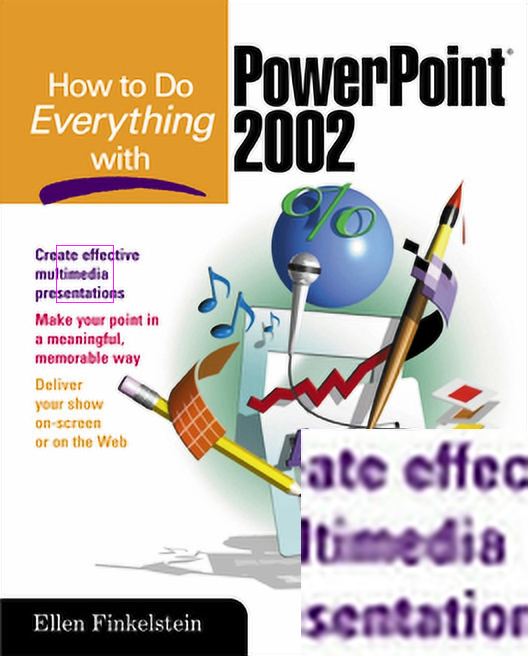}\label{baboon s-original}}
	\subfloat[][SRCNN MSCE 2x] {\includegraphics[width=3.1cm]{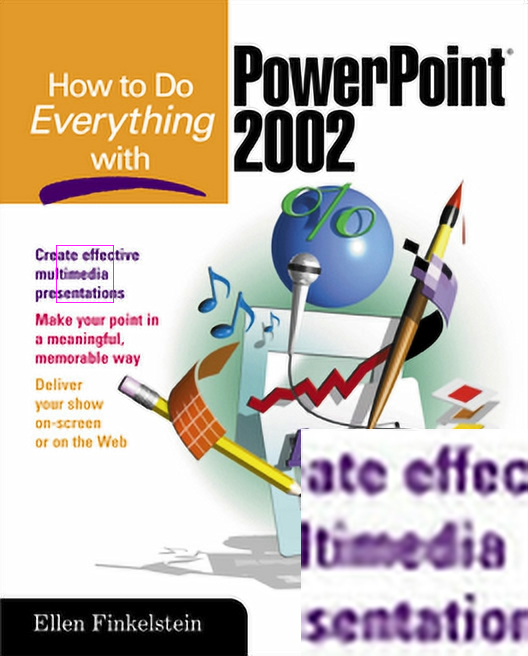}\label{baboon s-custom}}
	\subfloat[][ESPCN 2x] {\includegraphics[width=3.1cm]{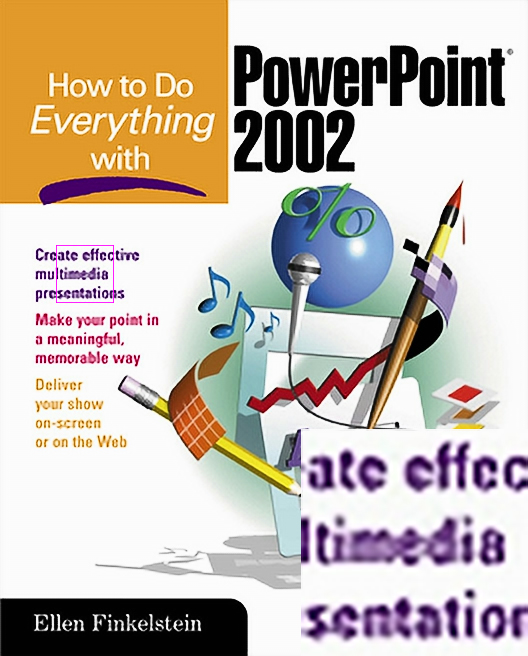}\label{baboon e-original}}
	\subfloat[][ESPCN MSCE 2x] {\includegraphics[width=3.1cm]{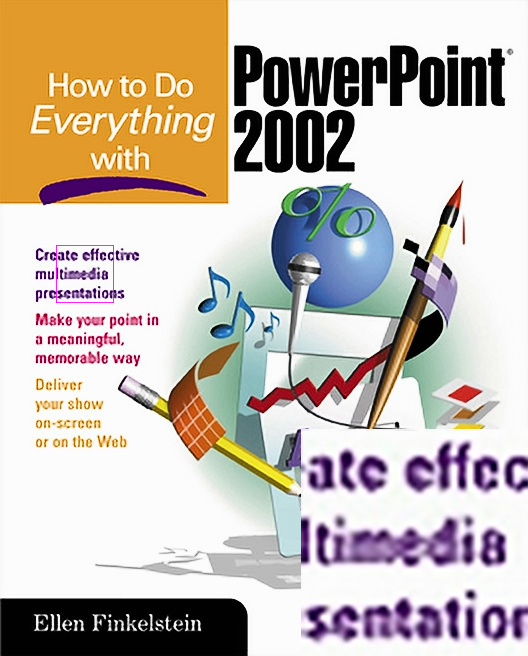}\label{baboon e-custom}}\\
    \subfloat[][SRCNN 3x] {\includegraphics[width=3.1cm]{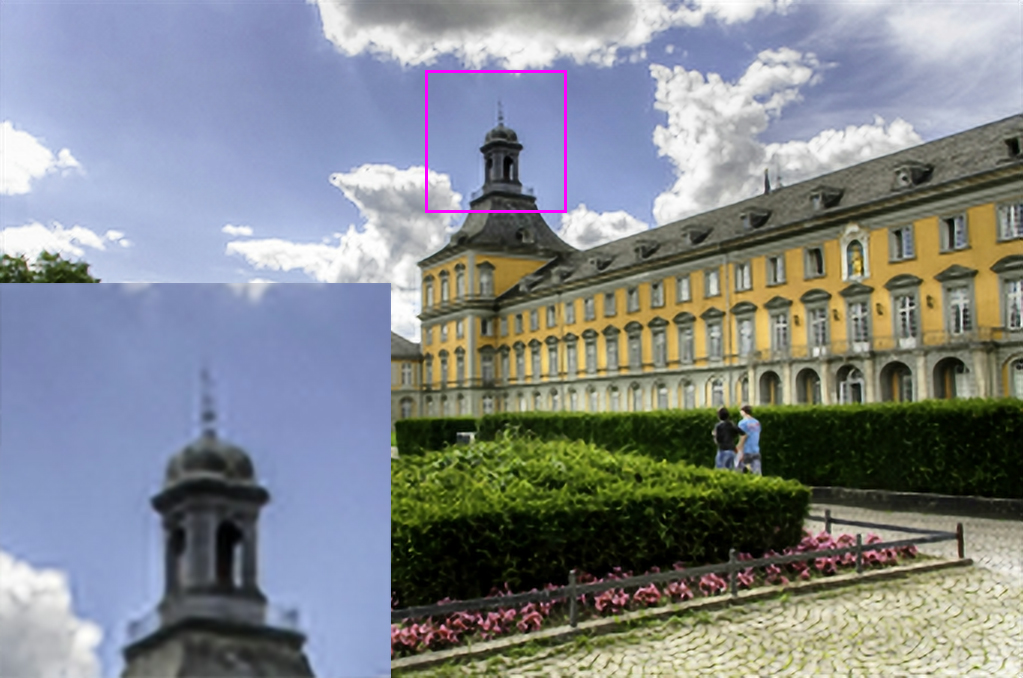}\label{baboon s-original}}
	\subfloat[][SRCNN MSCE 3x] {\includegraphics[width=3.1cm]{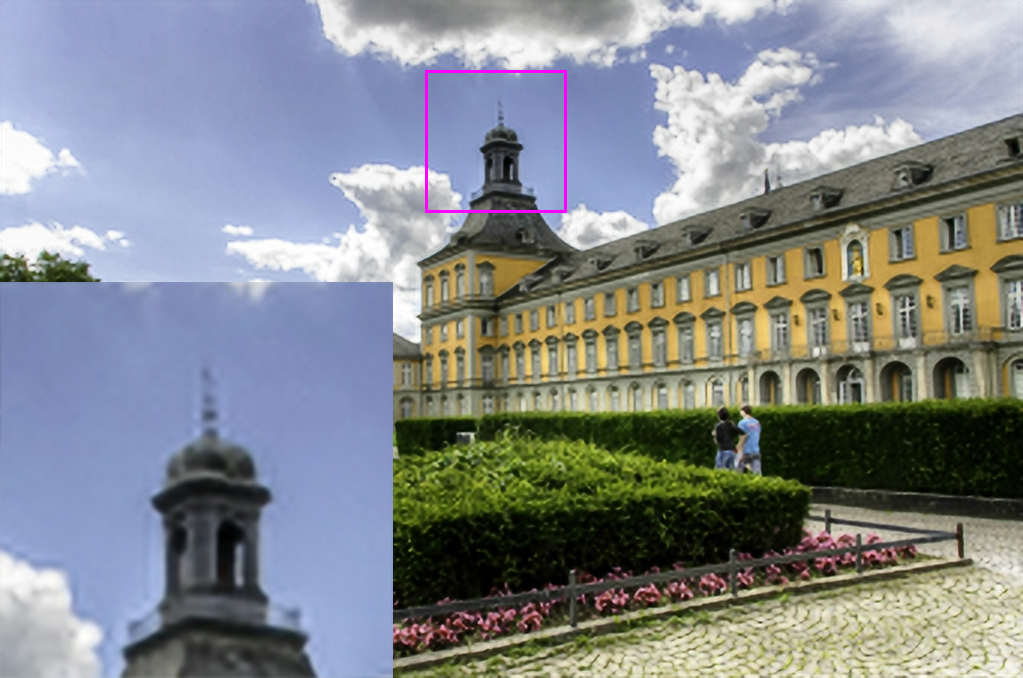}\label{baboon s-custom}}
	\subfloat[][ESPCN 3x] {\includegraphics[width=3.1cm]{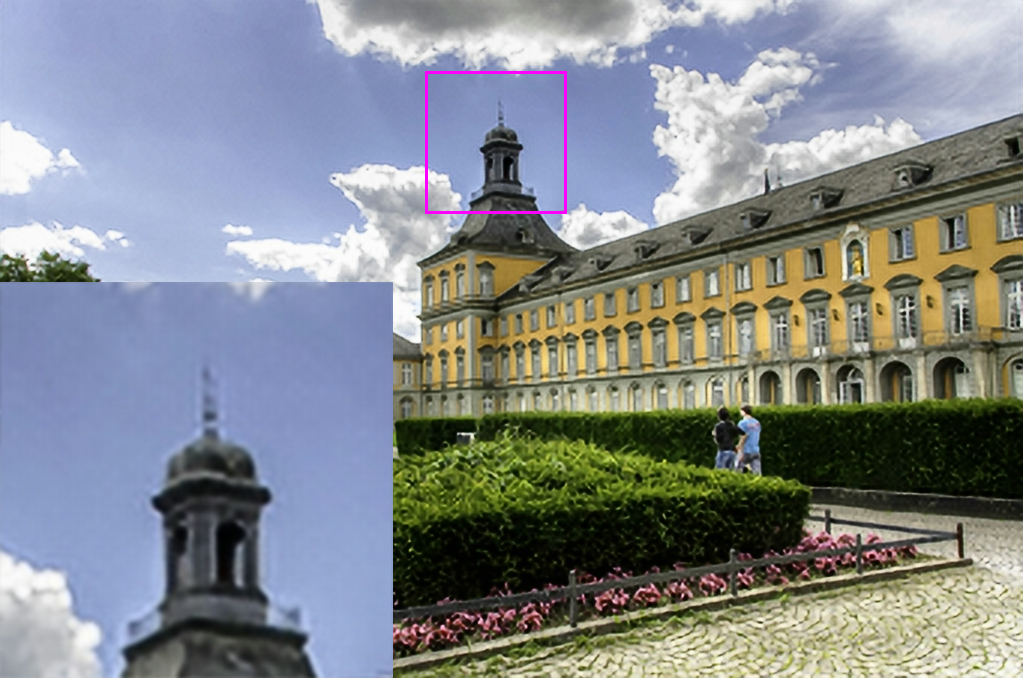}\label{baboon e-original}}
	\subfloat[][ESPCN MSCE 3x] {\includegraphics[width=3.1cm]{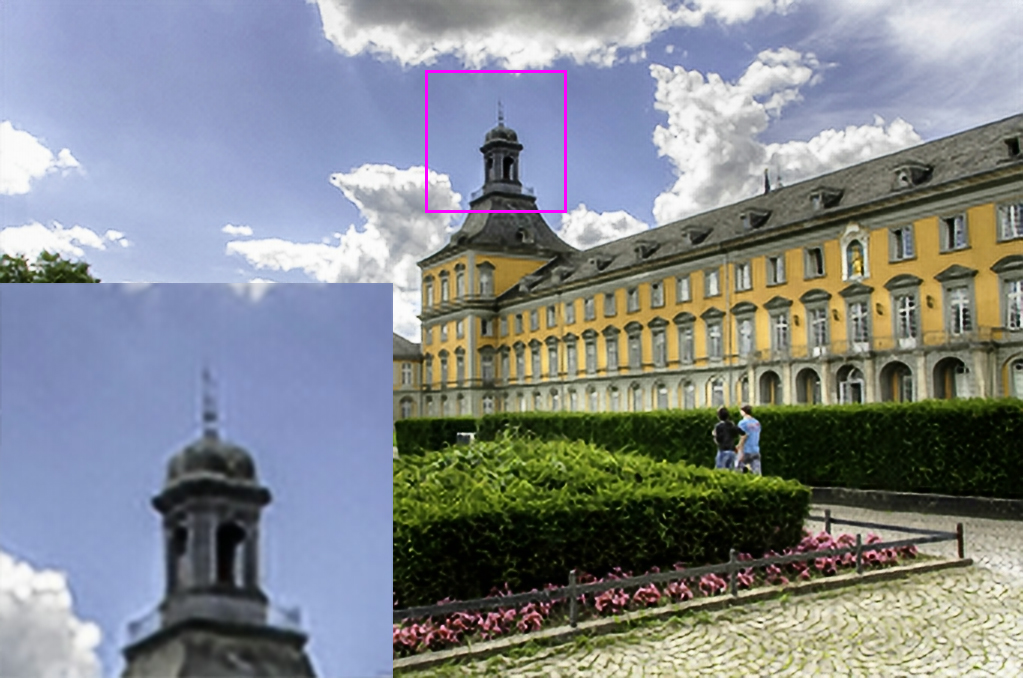}\label{baboon e-custom}}\\
	\subfloat[][SRCNN 4x] {\includegraphics[width=3.1cm]{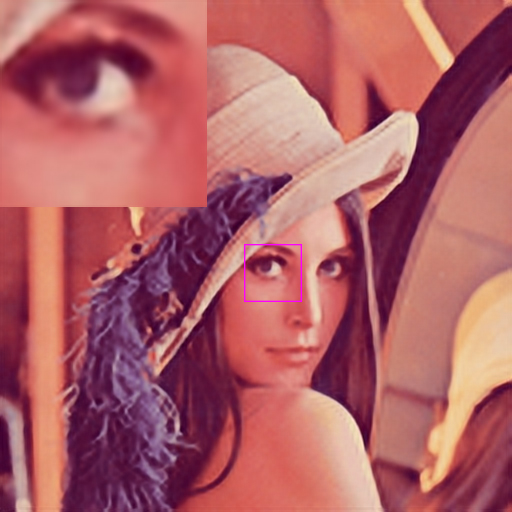}\label{baboon s-original}}
	\subfloat[][SRCNN MSCE 4x] {\includegraphics[width=3.1cm]{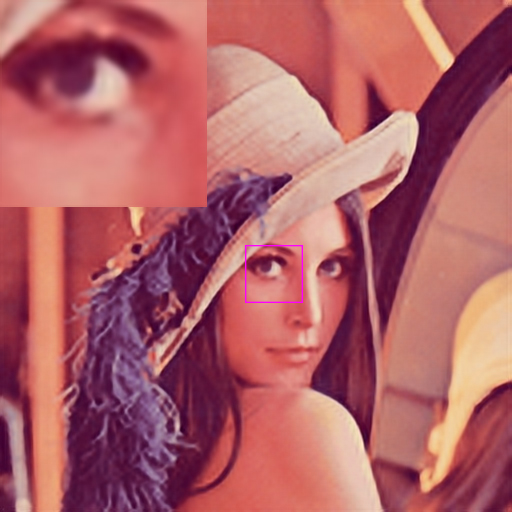}\label{baboon s-custom}}
	\subfloat[][ESPCN 4x] {\includegraphics[width=3.1cm]{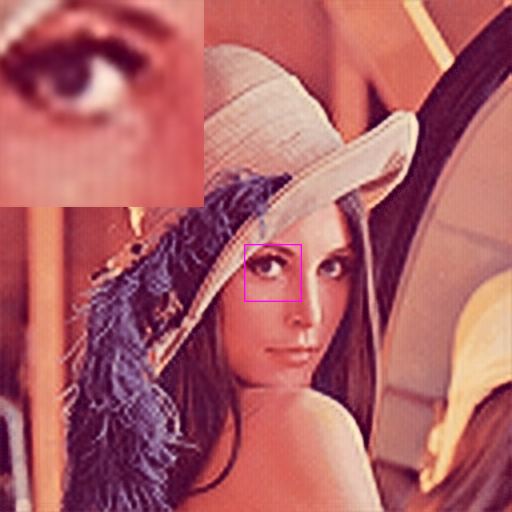}\label{baboon e-original}}
	\subfloat[][ESPCN MSCE 4x] {\includegraphics[width=3.1cm]{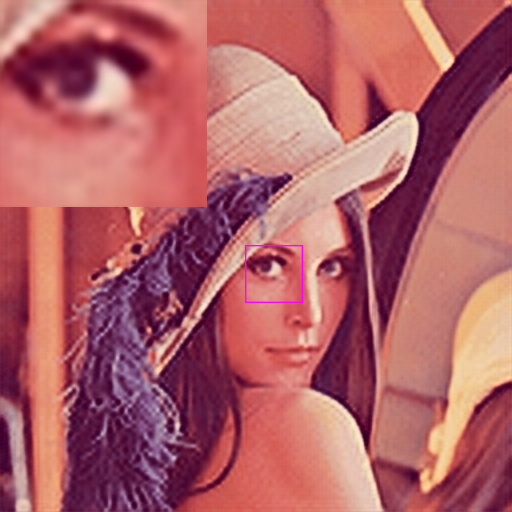}\label{baboon e-custom}}\\
	\subfloat[][SRCNN 8x] {\includegraphics[width=3.1cm]{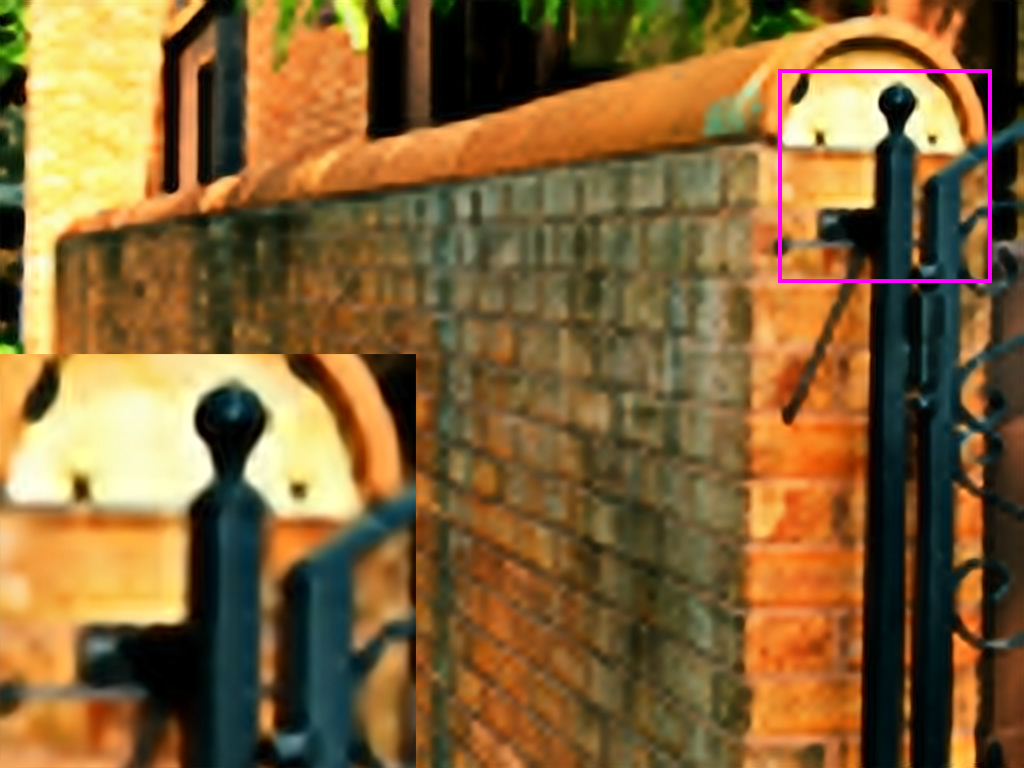}\label{baboon s-original}}
	\subfloat[][SRCNN MSCE 8x] {\includegraphics[width=3.1cm]{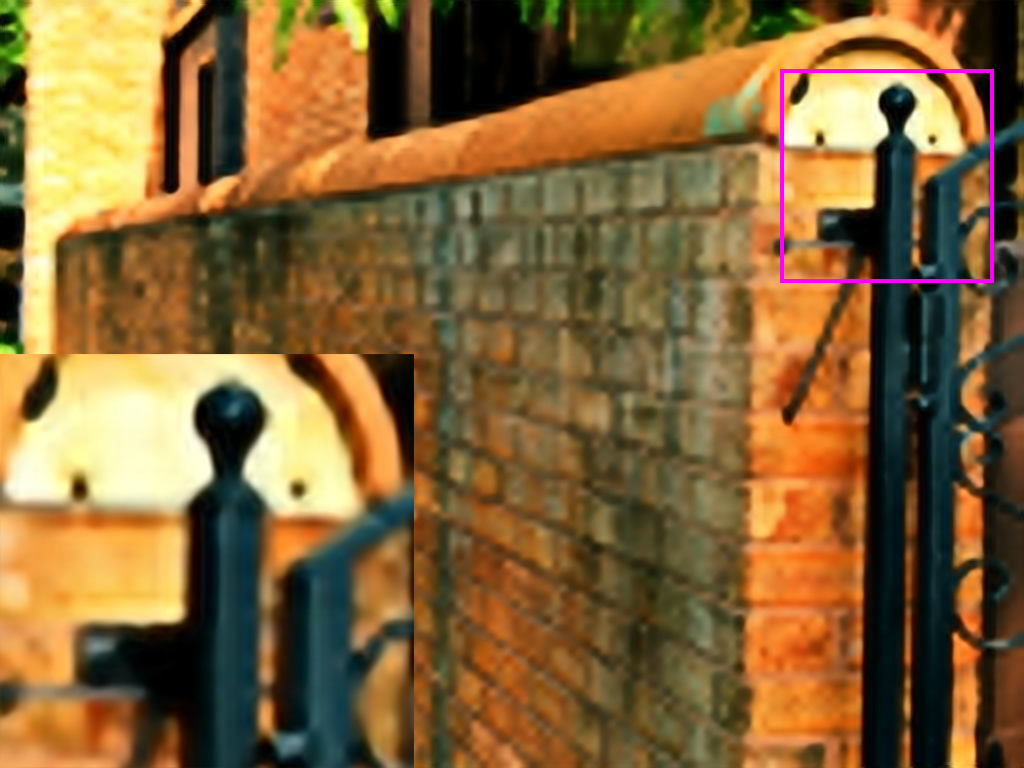}\label{baboon s-custom}}
	\subfloat[][ESPCN 8x] {\includegraphics[width=3.1cm]{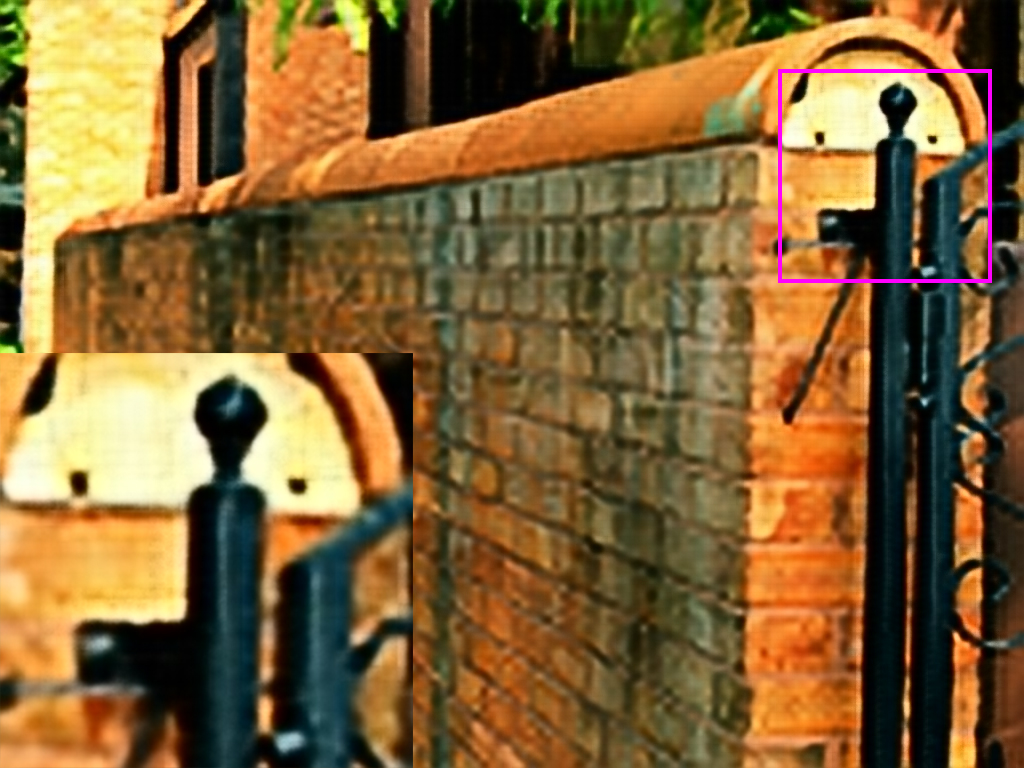}\label{baboon e-original}}
	\subfloat[][ESPCN MSCE 8x] {\includegraphics[width=3.1cm]{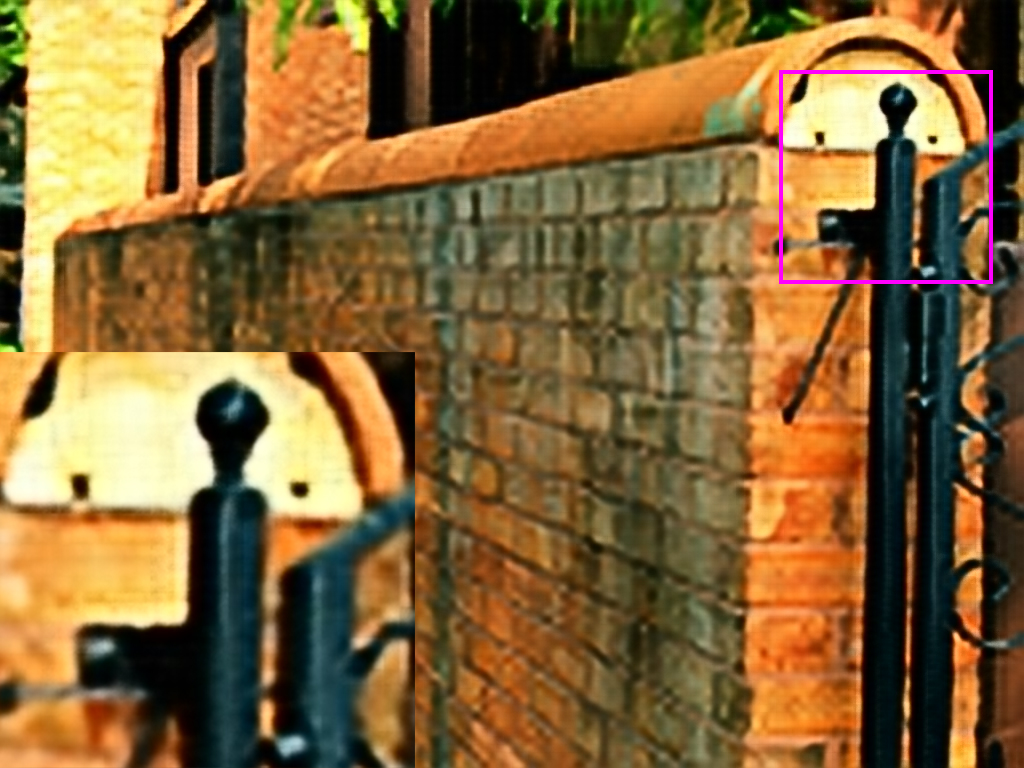}\label{baboon e-custom}}
        
	\caption{Comparison of the results obtained on down-sampled (by bicubic interpolation without blurring) images on different upscaling factors. (a), (b), (c) and (d) have been down-sampled by a factor of 2 and reconstructed. (e), (f), (g) and (h) have been down-sampled by a factor of 3 and reconstructed. (i), (j), (k) and (l) have been down-sampled by a factor of 4 and reconstructed. (m), (n), (o) and (p) have been down-sampled by a factor of 8 and reconstructed.}
	\label{Fig5}
\end{figure}
	
\begin{table*}[!]
	\centering
		
	\caption{$P_{*}$ and $S_{*}$ are the PSNR and SSIM values obtained by the method \(*\) for the upscaling factors of 2, 3, 4 and 8, whereas $P_{*}^{c}$ and $S_{*}^{c}$ are the corresponding PSNR and SSIM values obtained after augmenting the loss function by the MSCE loss function designed by us. All the models other than bicubic (non-learnable) have been trained on DIV2K training dataset. For testing, we have used 4 datasets, namely Set5, Set14, Urban and BSD.}

\begin{tabular}{| l  l | c  c | c c| c c| c c| c c|}
			
	\hline 
	\multicolumn{2}{|l |}{Dataset}   & $P_{bicubic}$ & $S_{bicubic}$ & $P_{srcnn}$&$S_{srcnn}$& $\bf P_{srcnn}^{c}$&$\bf S_{srcnn}^{c}$& $P_{espcn}$ &$S_{espcn}$ &$\bf P_{espcn}^{c}$ &  $ \bf S_{espcn}^{c}$ \\
			
	\hline
	\multirow{4}{*}{Set5}      & 2x &27.02 & 0.92& 28.44&0.93 &28.57 &0.93 &26.48 &0.92 &26.59 &0.92\\
			& 3x & 25.41 & 0.89 &26.59 &0.90 &26.75 &0.91  &25.882 &0.91 &25.888 &0.91\\
			& 4x & 21.96 & 0.79&23.22 &0.82 &23.37 &0.83  &22.35 &0.81 &22.49 &0.82\\
			& 8x & 18.10 & 0.61 &18.740 &0.63 &18.743 &0.63 &18.33 &0.62 &18.43 &0.62\\
	\hline 
	\multirow{4}*{Set14}       & 2x &24.10  & 0.86 &25.22  &0.88  &25.32 &0.88 &23.50 &0.87 &23.56 &0.87\\
			& 3x & 22.65 & 0.81 &23.62  & 0.84 &23.68 &0.84 &23.06 &0.84 &23.06 &0.84\\
			& 4x & 20.01 & 0.70 & 20.96 & 0.73 &21.04 &0.73 &20.12 &0.71 &20.32 &0.72\\
			& 8x & 17.13 & 0.53 & 17.57 & 0.56 &17.58 &0.56 &17.20 &0.56 & 17.27&0.56\\
	\hline 
	\multirow{4}*{Urban}      & 2x & 20.66 & 0.84 &22.26 &0.87 &22.44 &0.87 &21.38 &0.87 &21.42 &0.87\\
			& 3x & 20.22 & 0.79 &21.47 &0.83 &21.53 &0.83 &21.18 &0.83 &21.18 &0.83\\
			& 4x & 16.92 & 0.65 &17.81 &0.69 &17.84 &0.69 &17.54 &0.70 &17.59 &0.70\\
			& 8x & 14.63 & 0.48 &15.04 &0.50 &15.04 &0.50 &14.94 &0.507 &14.99 &0.509\\
	\hline 
	\multirow{4}*{BSD}       & 2x &25.88 & 0.89& 25.96&0.90 &26.18 &0.90 &23.36 &0.87 &23.41 &0.88\\
			& 3x & 21.86 & 0.77 &22.49 &0.81 &22.54 &0.81 &22.34 &0.81 &22.35 &0.81\\
			& 4x & 21.43 & 0.73&22.08 &0.77 &22.13 &0.77 &21.28 &0.76 &21.41 &0.77\\
			& 8x & 18.43 & 0.57 & 18.78&0.59 &18.81 &0.59 &18.47 &0.587 &18.58 &0.589\\
	\hline 
	\end{tabular}
	\label{result_table_all1}
		
\end{table*}
	
\begin{table*}[!]
	\centering
		
	\caption{$P_{*}$ and $S_{*}$ are the PSNR and SSIM values obtained by the method \(*\) at the different upscaling factors of 2, 3, 4 and 8, whereas $P_{*}^{c}$ and $S_{*}^{c}$ are the corresponding PSNR and SSIM values obtained by augmenting the loss function by the MSCE loss function designed by us. All the models other than bicubic (non-learnable) are trained on DIV2K (blurred by Gaussian blurring, then downsampled by bicubic) training dataset. For testing, we use 4 datasets, namely Set5, Set14, Urban and BSD.}

\begin{tabular}{| l  l | c  c | c c| c c| c c| c c|}
			
	\hline 
			\multicolumn{2}{|l |}{Dataset}   & $P_{bicubic}$ & $S_{bicubic}$ & $P_{srcnn}$&$S_{srcnn}$& $\bf P_{srcnn}^{c}$&$\bf S_{srcnn}^{c}$& $P_{espcn}$ &$S_{espcn}$ &$\bf P_{espcn}^{c}$ &  $ \bf S_{espcn}^{c}$ \\
			
	\hline
	\multirow{4}{*}{Set5}      & 2x &21.10 & 0.77& 23.96&0.83 &24.06 &0.84 &21.21 &0.75 &21.87 &0.79\\
			& 3x & 21.63& 0.79 &22.38 &0.85 &24.75 &0.86  & 22.50 & 0.80 & 22.88 & 0.83\\
			& 4x & 20.12 & 0.72&21.92 &0.78 &21.94 &0.78  &21.53 &0.77 &21.92 &0.78\\
			& 8x & 17.72 & 0.59 &18.17 &0.61 &18.34 &0.61 &18.48 &0.613 &18.56 &0.614\\
			\hline 
			\multirow{4}*{Set14}       & 2x &19.35  & 0.67 & 21.75  & 0.76  &21.78 &0.76 &19.50 &0.67 &20.08 &0.70\\
			& 3x & 19.84 & 0.69 &20.64  & 0.77 &22.25 &0.78 & 20.63&0.73 &20.99 &0.74\\
			& 4x & 18.67 & 0.62 & 20.02 & 0.69 &20.07 &0.69 &19.75 &0.68 &20.04 &0.69\\
			& 8x & 16.84 & 0.52 & 17.16 & 0.54 &17.29 &0.54 &17.35 &0.55 & 17.43&0.54\\
			\hline 
			\multirow{4}*{Urban}      & 2x & 16.57 & 0.63 &18.87 &0.74 &18.86 &0.74 &16.93 &0.63 &17.33 &0.66\\
			& 3x & 17.63 & 0.66 &18.96 &0.76 &20.03 &0.76 &18.68 &0.71 &18.87 &0.72\\
			& 4x & 15.85 & 0.58 &17.05 &0.65 &17.07 &0.65 &16.96 &0.64 &17.05 &0.64\\
			& 8x & 14.42 & 0.47 &14.74 &0.49 &14.81 &0.49 &14.96 &0.49 &14.97 &0.48\\
			\hline 
			\multirow{4}*{BSD}        & 2x &21.00 & 0.71& 23.27&0.80 &23.28 &0.80 &20.92 &0.71 &21.67 &0.75\\
			& 3x & 19.82 & 0.66 &20.27 &0.75 &21.59 &0.75 &20.30 &0.70 &20.60 &0.72\\
			& 4x & 20.13 & 0.67&21.32 &0.73 &21.35 &0.73 &21.06 &0.73 &21.38 &0.73\\
			& 8x & 18.18 & 0.56 & 18.38&0.58 &18.51 &0.58 &18.70 &0.58 &18.70 &0.58\\
			\hline 
			
		\end{tabular}
		
		\label{result_table_all2}
		
	\end{table*}

	\section{Conclusion}
	
	A large number of research papers have been published in the recent past by designing different models or algorithms that work reasonably well. The unique contribution of our work is that it improves the performance of any existing method, rather than proposing another technique. In this paper, we have proposed a robust edge-preserving loss function that adds performance gain in terms of PSNR and SSIM to any existing model, without increasing the computational cost involved in testing. We train the existing model by adding weighted Canny edge based loss. Minimizing this loss function helps to preserve the edges by giving more weightage to the edges. As shown by the Tables of results, the PSNR and SSIM values obtained after including our MSCE loss function are consistently better.

\pagebreak


\begin{thebibliography}{4}
\bibitem{glasner}
Glasner, Daniel, Shai Bagon, and Michal Irani.:Super-resolution from a single image. IEEE 12th Int. Conf. Comput. Vis., (2009).
\bibitem{srcnn14}
		Dong, C., Loy, C.C., He, K. and Tang, X.:Learning a deep convolutional network for image super-resolution. European Conf. Comput. Vis. Springer, Cham (2014).
\bibitem{subpixel}
Shi, Wenzhe, Jose Caballero, Ferenc Huszár, Johannes Totz, Andrew P. Aitken, Rob Bishop, Daniel Rueckert, and Zehan Wang.:Real-time single image and video super-resolution using an efficient sub-pixel convolutional neural network. Proc. IEEE Conf. Comput. Vis. and Pattern Recog.(2016).
 \bibitem{rkagrnatural}
Pandey, Ram Krishna, and A. G. Ramakrishnan.:A hybrid approach of interpolations and CNN to obtain super-resolution. arXiv preprint arXiv:1805.09400, (2018).
	\bibitem{rkagrdocument1}
    Pandey, Ram Krishna, and A. G. Ramakrishnan.:Language-independent single document image super-resolution using CNN for improved recognition. arXiv preprint arXiv:1701.08835, (2017).
\bibitem{rkagrdocument2}
Pandey, Ram Krishna, and A. G. Ramakrishnan.:Efficient document-image super-resolution using convolutional neural network. Sadhana 43:15, March (2018).
\bibitem{rkagrdocument3}
Pandey, Ram Krishna, Shishira R. Maiya, and A. G. Ramakrishnan.:A new approach for upscaling document images for improving their quality. Proc. 14th IEEE INDICON 2017, IIT Roorkee, Dec. 15-17, (2017).
	\bibitem{srgan}
Ledig, C., Theis, L., Huszár, F., Caballero, J., Cunningham, A., Acosta, A., Aitken, A., Tejani, A., Totz, J., Wang, Z. and Shi, W.:Photo-realistic single image super-resolution using a generative adversarial network. Proc. IEEE Conf. Comput. Vis. and Pattern Recog., (2017).
	\bibitem{perceptualloss}
J. Johnson, A. Alahi, and F. Li.:Perceptual losses for real-time style transfer and super-resolution. In European Conf. Comput. Vis. (ECCV), pages 694-–711. Springer, (2016).

\bibitem{vggnet}
K. Simonyan and A. Zisserman.:Very deep convolutional networks for large-scale image recognition. In Int. Conf. Learning Representations (ICLR), (2015).
	\bibitem{canny}
	J. Canny.:A Computational approach to edge detection. IEEE Trans. on Pattern Analysis and Machine Intelligence, vol. PAMI-8, no. 6, pp. 679-698, Nov. 1986. doi: 10.1109/TPAMI.1986.4767851
\bibitem{featurevisualization}
Simonyan, K., Vedaldi, A., Zisserman, A.: Deep inside convolutional net-works: visualising image classification models and saliency maps. arXiv preprint arXiv:1312.6034, (2013).
\bibitem{Gatystexture}
Gatys, L.A., Ecker, A.S., Bethge, M.: Texture synthesis using convolutional neural networks. In: Advances in Neural Information Processing Systems 28, May (2015).
\bibitem{Gatysstyle}
Gatys, L.A., Ecker, A.S., Bethge, M.: A neural algorithm of artistic style. arXiv preprint arXiv:1508.06576, (2015).
\bibitem{Johnson}
Johnson, Justin, Alexandre Alahi, and Li Fei-Fei. :Perceptual losses for real-time style transfer and super-resolution. European Conference on Computer Vision. Springer, Cham, (2016).
\bibitem{rkagrart}
 Pandey, Ram Krishna, Samarjit Karmakar, and A. G. Ramakrishnan.:Computationally efficient approaches for image style transfer. arXiv preprint arXiv:1807.05927, (2018).

\bibitem{adam}
Kinga, D., and J. Ba.:Adam: A method for stochastic optimization. Proc. Int. Conf. Learning Representations (ICLR), (2015).
\bibitem{div2k}
Eirikur Agustsson and Radu Timofte.:NTIRE 2017 challenge on single image super-resolution: dataset and study. Proc. IEEE Conf. Comput. Vis. and Pattern Recog. (CVPR) Workshops, July (2017).

\bibitem{set5}
Marco Bevilacqua, Aline Roumy, Christine Guillemot and Marie-Line A Morel.:Low-complexity single-image super-resolution based on nonnegative neighbor embedding. Proc. BMVC (2012).
\bibitem{set14}
R Zeyde, M. Elad and M. Protter.:On single image scale-up using sparse-representations. Proc. Int. Conf. Curves and Surfaces, Avignon-France, June 24-30, (2010).
		
\bibitem{bsd100}
D. Martin, C. Fowlkes, D. Tal and J. Malik.:A database of human segmented natural images and its application to evaluating segmentation algorithms and measuring ecological statistics. Proc. 8th Int. Conf. Comput. Vis., vol. 2, 416--423, July (2001).
\bibitem{urban100}
Jia-Bin Huang, Abhishek Singh and Narendra Ahuja.:Single image super-resolution from transformed self-exemplars.  Proc. IEEE Conf. Comput. Vis. and Pattern Recog., 5197--5206, (2015).

\end{thebibliography}
\end{document}